\documentclass{article}

\usepackage{arxiv}

\usepackage[utf8]{inputenc} 
\usepackage[T1]{fontenc}    
\usepackage{graphicx}       
\usepackage{hyperref}       
\usepackage{url}            
\usepackage{booktabs}       
\usepackage{nicefrac}       
\usepackage{microtype}      
\usepackage{xcolor}         
\usepackage{amsfonts}       
\usepackage{amsmath}
\usepackage{bbm}
\usepackage[comma,semicolon]{natbib}
\usepackage{doi}

\newcommand{\norm}[1]{\left\lVert#1\right\rVert}

\title{Continual Learning and Catastrophic Forgetting}
\author{
	\href{https://orcid.org/0000-0002-5239-5660}{\includegraphics[scale=0.06]{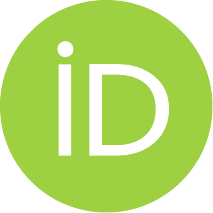}\hspace{1mm}Gido M.~van de Ven}$^*$ \\
	Department of Electrical Engineering\\
	KU Leuven, Belgium\\
	\texttt{gido.vandeven@kuleuven.be} \\
\And
    Nicholas Soures\thanks{Joint first author. This preprint is distributed under a read-only license.} \\
	Department of Electrical and Computer Engineering\\
	University of Texas at San Antonio, USA\\
	\texttt{nms9121@rit.edu} \\
\And
   \href{https://orcid.org/0000-0003-4462-5224}{\includegraphics[scale=0.06]{figures/orcid.pdf}\hspace{1mm}Dhireesha Kudithipudi} \\
   Department of Electrical and Computer Engineering\\
   University of Texas at San Antonio, USA\\
   \texttt{dk@utsa.edu} \\
}
\date{}

\renewcommand{\headeright}{}
\renewcommand{\undertitle}{}
\renewcommand{\shorttitle}{Continual Learning and Catastrophic Forgetting}

\hypersetup{
    pdftitle={Continual Learning and Catastrophic Forgetting},
    pdfsubject={cs.LG, cs.AI, q-bio.NC, q-bio.QM},
    pdfauthor={Gido M van de Ven, Nicholas Soures, Dhireesha Kudithipudi},
    pdfkeywords={Continual learning, Catastrophic forgetting, Deep learning, Cognitive science, Incremental learning, Lifelong learning},
}

\begin{document}
\maketitle

\vspace{-5pt}
\begin{abstract}
This book chapter delves into the dynamics of continual learning, which is the process of incrementally learning from a non-stationary stream of data. Although continual learning is a natural skill for the human brain, it is very challenging for artificial neural networks. An important reason is that, when learning something new, these networks tend to quickly and drastically forget what they had learned before, a phenomenon known as catastrophic forgetting. Especially in the last decade, continual learning has become an extensively studied topic in deep learning. This book chapter reviews the insights that this field has generated.
\end{abstract}


\section*{Key Points}
\begin{itemize}
    \item Incrementally learning from a non-stationary stream of data, referred to as continual learning, is a key aspect of intelligence.
    \item Artificial neural networks tend to rapidly and drastically forget previously learned information when learning something new, a phenomenon referred to as catastrophic forgetting.
    \item Catastrophic forgetting is an important reason why continual learning is so challenging for deep neural networks, but solving the continual learning problem requires more than preventing catastrophic forgetting.
    \item Two distinctions often made in the deep learning literature on continual learning are between task-based and task-free continual learning, and between task-, domain- and class-incremental learning. These two distinctions are orthogonal to each other, and can be captured in a single framework.
    \item Deep learning methods for continual learning are evaluated using metrics covering performance, diagnostic analysis and resource efficiency.
    \item Six main computational approaches for continual learning with deep neural networks are (i)~replay, (ii)~parameter regularization, (iii)~functional regularization, (iv)~optimization-based approaches, (v)~context-dependent processing and (vi)~template-based classification.
    \item Establishing further connections between deep learning and cognitive science in the context of continual learning could benefit both fields.
\end{itemize}


\section{Introduction}

Continual learning is a key aspect of intelligence. The ability to accumulate knowledge by incrementally learning from one’s experiences is an important skill for any agent, natural or artificial, that operates in a non-stationary world. Humans are excellent continual learners. The human brain can incrementally learn new skills without compromising those that were learned before, and it is able to integrate and contrast new information with previously acquired knowledge \citep{flesch2018comparing,kudithipudi2022biological}. Intriguingly, artificial deep neural networks, although rivaling human intelligence in other ways, almost completely lack this ability to learn continually.
Most strikingly, when these networks are trained on something new, they tend to ``catastrophically'' forget what they had learned before \citep{mccloskey1989catastrophic,ratcliff1990connectionist}.

The inability of deep neural networks to continually learn has important practical implications. Due to its powerful representation learning capabilities, deep learning has become a major driving force behind many recent advances in artificial intelligence. However, to achieve their strong performance, deep neural networks must be trained for extended periods of time on large amounts of data (e.g., \citealp{radford2021learning}). This resource-intensive training makes the lack of continual learning abilities of these networks particularly costly. For example, if after an expensive training process has finished, relevant new data become available, rapidly updating the network by training only on the new data does not work. To avoid adapting too strongly to the new data, the network must be trained on both old and new data together. However, even such continued joint training often does not yield satisfactory results. Instead, practitioners in industry tend to periodically retrain the entire network from scratch on all data, despite the large computational costs \citep{huyen2022}. Therefore, developing successful continual learning methods for deep learning could result in significant efficiency gains and a substantial reduction in the required resources. Another important potential application for continual learning is correcting errors or biases. After being trained, deep neural networks are often found to make mistakes or have certain biases (e.g., a subset of demographics is underrepresented in the training set), but updating a network to correct for these is difficult \citep{mitchell2022fast}.
Another practical use case of continual learning is in edge applications \citep{deng2020edge}. These applications require the ability to learn in real-time on-device to reduce their reliance on pre-deployed solutions.
For these and other reasons (e.g., see \citealp{verwimp2023continual}), continual learning has become an intensively studied topic in deep learning, and is seen as one of the main open challenges in the field \citep{hadsell2020embracing,bubeck2023sparks}.

In addition to the practical arguments above, another motivation for studying continual learning in deep neural networks is to gain insight into the computational principles that might underlie the cognitive processes supporting continual learning in the brain. Artificial deep neural networks are a popular class of computational models that can account for many aspects of information processing in the brain (e.g.,~\citealp{vangerven2017artificial,perconti2020deep,doerig2023neuroconnectionist}). Yet, in terms of continual learning, this class of models has clear insufficiencies. If we can understand the reasons for this failure and how to fix it, this might give clues as to which computational processes underlie the cognitive skill of continual learning. 
At the same time, but in the reverse direction, the brain's exceptional ability to continually learn can serve as a source of inspiration for the development of novel continual learning algorithms for deep learning.

The structure of this book chapter is as follows. Section~\ref{sec:problem} introduces the problem of continual learning, illustrates why it is so challenging for deep neural networks, and discusses different variants of the continual learning problem. Section~\ref{sec:approaches} covers various computational strategies that have been proposed to improve the continual learning capabilities of neural networks. Section~\ref{sec:humans} compares how continual learning is studied in cognitive science versus in deep learning, and section~\ref{sec:conclusion} concludes.


\section{The Continual Learning Problem}
\label{sec:problem}

Continual learning is the skill of \emph{incrementally} learning from a \emph{non-stationary} stream of data. In this definition, the term ``non-stationary'' indicates that the distribution of the data from which is learned changes over time. The term ``incrementally'' signals that new learning should not overwrite what was learned before, but that knowledge should be accumulated.
In this section, we discuss why continual learning is so challenging for artificial deep neural networks (subsections~\ref{sec:catastrophic_forgetting} and~\ref{sec:other_features}), we review different types of continual learning (subsections~\ref{sec:task_free} and~\ref{sec:scenarios}), and we briefly discuss how continual learning is evaluated (subsection~\ref{sec:evaluation}).


\subsection{Catastrophic Forgetting}
\label{sec:catastrophic_forgetting}

\begin{figure*}[t]
\begin{center}
\includegraphics[width=0.54\linewidth]{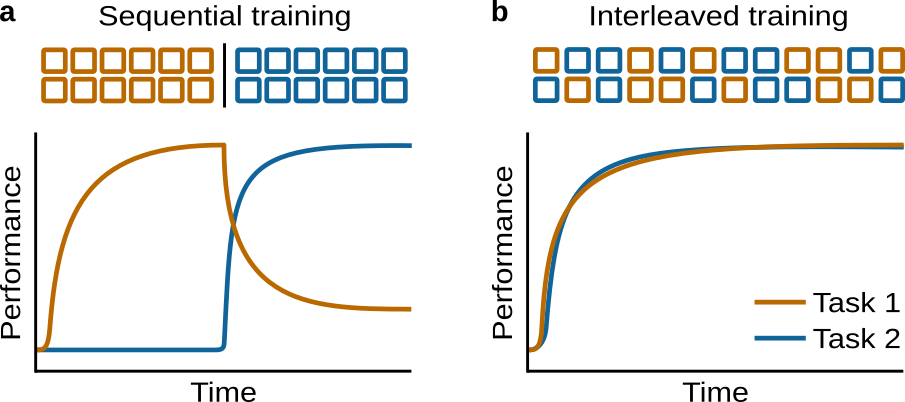}
\caption{\label{fig:continuity}\textbf{Schematic illustration of catastrophic forgetting.} \textbf{a},~When an artificial deep neural network is sequentially trained on two tasks, it rapidly and drastically forgets the first task while training on the second one. \textbf{b},~Importantly, when instead trained in an interleaved fashion, the same network is able to learn both tasks, illustrating that catastrophic forgetting is not due to limited model capacity.} 
\end{center}
\end{figure*}

Central to continual learning is the concept of catastrophic forgetting, also referred to as catastrophic interference.
Catastrophic forgetting is the phenomenon that artificial neural networks tend to rapidly and drastically forget previously learned information when learning new information (see Fig.~\ref{fig:continuity}).

When a neural network is sequentially trained on multiple tasks, forgetting of earlier tasks can be expected because the network parameters are adjusted to optimize the loss on the new task, which likely pushes these parameters away from their optimum value that was found for the earlier tasks. It was appreciated early on that such forgetting would occur when incrementally training a neural network on multiple tasks, but initially it was speculated that this forgetting might be relatively mild: \citet{hinton1986distributed} hypothesized that the many small parameter updates that work together to optimize the new task, might mostly cancel each other out in terms of their effect on previous tasks. This however turned out not to be the case.
\citet{mccloskey1989catastrophic} and \citet{ratcliff1990connectionist} were the first to demonstrate that the sequential training of simple neural networks on disjoint sets of data results in drastic forgetting, even with only small amounts of training on the new data distribution. \citeauthor{mccloskey1989catastrophic} also noted that this forgetting is substantially worse than that observed in humans, which prompted them to describe it as ``catastrophic''. 


Inspired by the observations from \citeauthor{mccloskey1989catastrophic} and \citeauthor{ratcliff1990connectionist}, in the 1990s several research groups began exploring the problem of catastrophic forgetting in early connectionist models. The proposed solutions incorporated concepts such as reducing network overlap (e.g.,~sparse or orthogonal representations) and the rehearsal of prior data (see \citealp{robins1995catastrophic} and \citealp{french1999catastrophic} for reviews). Following these works, the early concepts of continual learning began to emerge \citep{thrun1995lifelong}. More recently, \citet{srivastava2013compete} and \citet{goodfellow2013empirical} demonstrated that catastrophic forgetting remains a problem in modern deep learning architectures, which had just begun gaining popularity. Since then there has been a rapid rise in the number of deep learning studies addressing the problem of continual learning.

\subsection{Other Features Important for Continual Learning}
\label{sec:other_features}

While catastrophic forgetting is the root cause for why continual learning is so challenging for neural networks, it is not the case that preventing catastrophic forgetting is enough to solve the continual learning problem. One reason for this is that approaches that lower forgetting often introduce or exacerbate other issues. For example, reduced forgetting often comes at the cost of an impaired ability to learn new information, a trade-off known as the stability-plasticity dilemma \citep{grossberg1982processing}. Another example is that catastrophic forgetting could be avoided by always training on all data seen so far, but this approach presents significant challenges in terms of memory and computation usage.
Among others, \cite{diaz2018don}, \cite{mundt2022clevacompass}, \cite{prabhu2023computationally} and \cite{verwimp2023continual} have stressed that continual learning should focus on more than just preventing forgetting.

In a recent perspective article, \citet{kudithipudi2022biological} argued that, in addition to avoiding catastrophic forgetting, successful continual learning methods should exhibit the following features for full-scale operation:

\begin{enumerate}
   \item \textbf{Adaptation} - 
   Continual learning models must be able to quickly adapt to new situations or surroundings, without requiring extensive offline (re)training. Such rapid adaptation is essential when models are deployed in the real world, where conditions may vary considerably and change quickly.
   It was recently shown that incremental training of deep neural networks on a sequence of tasks can lead to a substantial loss of plasticity \citep{dohare2023loss}, highlighting that rapid adaptation is still an open problem in continual learning.

   \item \textbf{Exploiting task similarity} -
   When the different tasks (or contexts) that must be learned are related, it should be possible to exploit their similarity to achieve `positive transfer' between tasks. Positive transfer means that due to learning one task, the network also becomes better at another task~--~either directly in terms of performance improvement, or indirectly by making (re)learning of that task easier. For example, once a human has learned to play a first musical instrument (e.g.,~the piano), it is typically easier for them to master a second one (e.g.,~the violin). In continual learning, there are two types of transfer: \emph{forward transfer}, whereby learning a new task facilitates future tasks; and \emph{backward transfer}, whereby learning a new task benefits previously learned tasks. Especially positive backward transfer has proven to be a challenging feature to achieve with artificial deep neural networks \citep{lopez2017gradient}. A promising route to enabling improved knowledge transfer between tasks is learning compositional representations \citep{mendez2023how}.
 
    \item \textbf{Task agnostic} -
    In many real world problem settings, continual learning models cannot rely on an oracle to tell them for each encountered example to which task (or domain / context) it belongs. It is often a desirable property for continual learning models to be \emph{task agnostic}. The term task agnostic can refer to multiple different things: not knowing task identity during testing, not knowing task identity during training, not being informed about task switches, or even that there is no discrete set of underlying tasks at all. Various forms of being task agnostic are covered in subsections~\ref{sec:task_free} and~\ref{sec:scenarios}.

    \item \textbf{Noise tolerance} -
    Deep learning models are usually trained on datasets that are curated, cleaned, and annotated to optimize training (e.g., ImageNet; \citealp{deng2009imagenet}). Training instead on non-curated datasets can lead to substantially lower performance and generalization capabilities \citep{tian2021divide}. For practical applications of continual learning, it is important that models are able to deal with data in the form they arrive~--~raw, noisy and uncleaned; as well as with situations in which the noise level or distribution changes over time, for example due to variability in the environment or in the sensors.

    \item \textbf{Resource efficiency and sustainability} -
    When deep neural networks are expected to continue learning for extended periods of time, it is critical that they do so in a sustainable and resource efficient manner. Existing continual learning methods often violate this desideratum \citep{vogelstein2020omnidirectional,prabhu2023computationally}.
    For example, storing all data points that are encountered (e.g., in a memory buffer) can be problematic in terms of required storage space, while constantly retraining on all previous tasks can quickly become infeasible in terms of computational costs.
    \citet{verwimp2023continual} argued that he goal of continual learning can be interpreted as finding the best approach in terms of some trade-off between performance and resource efficiency (e.g.,~memory and compute), whereby the relative importances in the trade-off differ depending on the problem.

\end{enumerate}


\subsection{Task-based versus Task-free Continual Learning}
\label{sec:task_free}

\begin{figure*}[b]
\begin{center}
\vspace{2pt}
\includegraphics[width=0.98\linewidth]{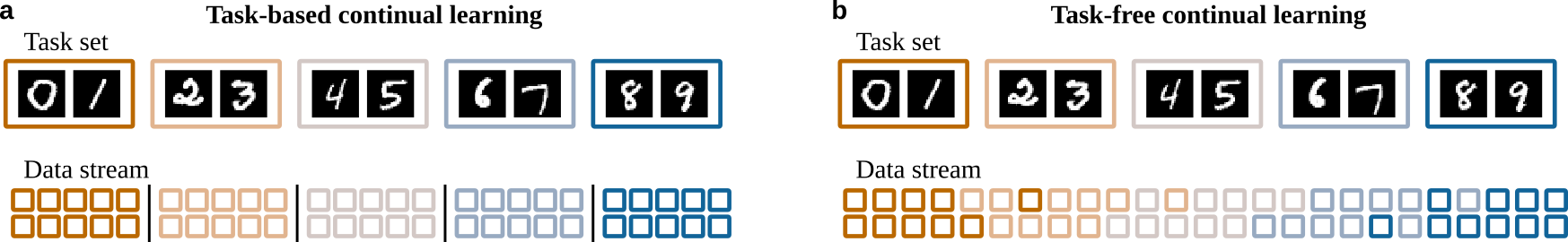}
\caption{\label{fig:task_based}\textbf{Task-based and task-free continual learning illustrated with Split MNIST.} \textbf{a},~In standard task-based continual learning, there is a discrete set of tasks that are presented sequentially, often with marcated boundaries between tasks. This is illustrated here with Split MNIST, where the MNIST dataset is split up into five tasks so that each task contains two digits. \textbf{b},~In task-free continual learning, there is typically still a discrete set of underlying tasks, but the transitions between tasks are continuous. In the Split MNIST example, the probabilities of observing the examples of each digit gradually change over time.}
\end{center}
\end{figure*}

A common assumption in continual learning research is that there is a discrete set of tasks that are presented to the network one after the other, often with marked boundaries between tasks. This \emph{task-based continual learning} setting is illustrated in Fig.~\ref{fig:task_based}a.
A popular way to set up a task set for continual learning is to take an existing dataset (e.g.,~MNIST, \citealp{lecun1998gradient}; CIFAR-100, \citealp{krizhevsky2009learning}; MiniImageNet, \citealp{vinyals2016matching}) and split it into tasks based on the class labels (e.g.,~Split MNIST; see Fig.~\ref{fig:task_based}). Another common way to create multiple tasks from a single dataset is to use task-specific transformations, for example rotations, permutations or the addition of noise. An alternative approach is to interweave multiple datasets (e.g.,~one task contains the ten MNIST-digits, another task contains the ten digits from the SVHN dataset; \citealp{netzer2011reading}). Task-based continual learning is an important and frequently used setting in the literature because it provides a convenient way to study various aspects of continual learning in a controlled and isolated manner.

However, not all aspects of the rich pattern of non-stationarity in the real world are well captured by the task-based continual learning setting. For example, continual learning methods developed in the task-based setting can be reliant on the presence of hard boundaries between tasks (to perform certain consolidation operations, such as updating the memory buffer or replacing a stored copy of the model) \citep{aljundi2019task,zeno2019task}, and for many methods it is unclear how they could benefit if a previously seen task is encountered again \citep{stojanov2019incremental,hemati2023class}.
This has motivated the emergence of \emph{task-free continual learning} (Fig.~\ref{fig:task_based}b), which allows for gradual transitions between tasks and repetition of tasks. With task-free continual learning, there is usually still an underlying set of discrete tasks (e.g., \citealp{Lee2020A,jin2021gradient,shanahan2021encoders}), but the transitions between those tasks are continuous, in the sense that the probabilities of observing each task gradually change over time.
A description of how the probabilities of observing each task change over time has been referred to as a \emph{schedule} \citep{shanahan2021encoders,wang2022schedule}.

Following \citet{van2022three}, the above descriptions of task-based and task-free continual learning can be formalized by letting $\mathcal{T}$ be the set of underlying tasks and defining the data stream as a sequence of experiences: $\{e_1, e_2, ..., e_N\}$. These experiences $e_n$ are the `incremental steps' of a continual learning problem, in the sense that they are presented one after the other and the network has free access to the data of the current experience but not to the data from other experiences (except possibly to data stored in a memory buffer). In the standard task-based continual learning setting, each experience contains all training data for its corresponding task: $e_t=\mathcal{D}^{\text{train}}_t$ for all $t\in\mathcal{T}$. In the task-free continual learning setting, each data point within every experience can be sampled from any combination of underlying tasks:
\begin{equation}
e_{n}[i]\sim\sum_{t\in\mathcal{T}}\pi^{n,i}_t\mathcal{D}_t
\label{eq:sampling}
\end{equation}
whereby $e_{n}[i]$ is data point~$i$ of experience~$n$ and $\pi^{n,i}_t$ is the probability that this data point is sampled from~$\mathcal{D}_t$, the data distribution of task~$t$. In this notation, the schedule is given by $\{\pi^{n,i}_t\}_{n\in\mathbb{N}_{\leq N},i\in\mathbb{N}_{\leq I_n},t\in\mathcal{T}}$ , where $I_n$ denotes the number of data points in experience~$n$. In this framework for continual learning, the task set~$\mathcal{T}$ describes \emph{what part} of the data change over time (i.e., the non-stationary aspect of the data) and the schedule~$\pi$ describes \emph{how} it changes over time (i.e., the temporal correlation structure). Finally, we note that this framework can be generalized to allow for a continuous task set (this requires changing the summation in equation~\ref{eq:sampling} to an integral), but as far as we are aware this option has not yet been systematically explored in the literature.

In addition to `task-based' and `task-free', other labels that have been used to describe variants of continual learning are `streaming' and `online'. Although the precise definitions of these terms vary, \emph{streaming continual learning} generally means that only a single training example is presented to the network at a time (i.e., in the above framework, each experience $e_n$ contains only one data point) \citep{hayes2020remind,banerjee2021class}, and \emph{online continual learning} typically refers to that the network encounters each training sample only once, also referred to as the `single-pass-through-data' setting \citep{aljundi2019online,chen2020mitigating}.


\subsection{Three Continual Learning Scenarios}
\label{sec:scenarios}

\begin{figure}[t]
  \centering
  \includegraphics[width=0.85\linewidth]{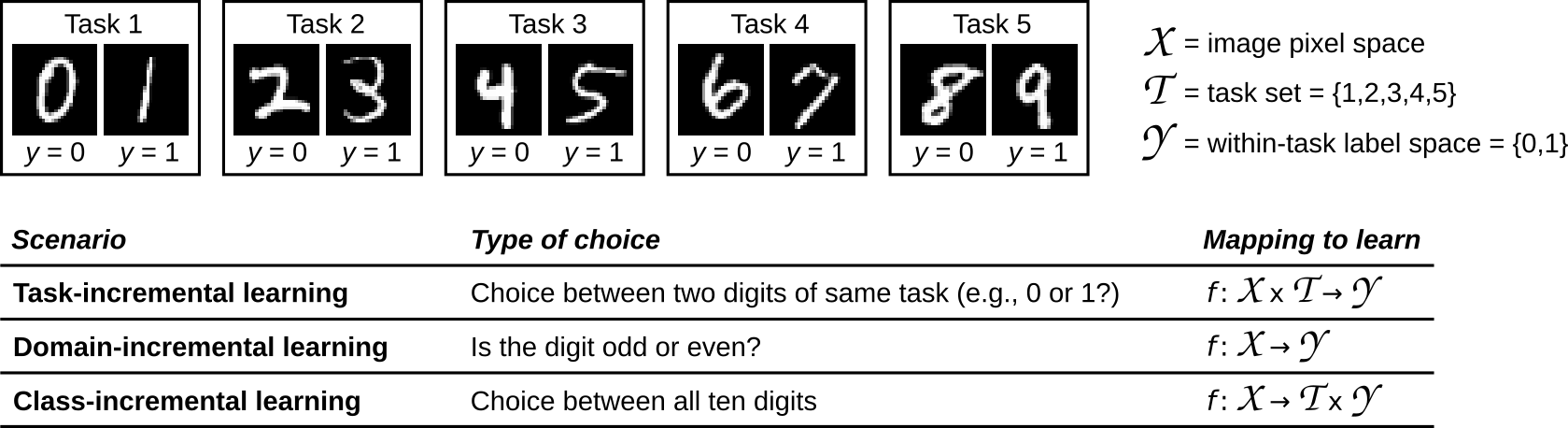}
  \caption{\textbf{Split MNIST according to each of the three continual learning scenarios.}
  With task-incremental learning, the choice for the network is between two digits from a known task (e.g., zero or one). With domain-incremental learning, the choice is still between two options (e.g., even or odd), but the problem is more difficult as task identity is not provided. With class-incremental learning, the network must choose between all ten digits. Modified from Fig.~2 of \citet{van2022three}.}
  \label{fig:scenarios}
\end{figure}

Another important way in which continual learning problems can differ from each other is described by \citet{van2018three}.
They distinguished three scenarios that have been widely used in the literature: task-incremental learning (sometimes abbreviated as Task-IL or TIL), domain-incremental learning (Domain-IL or DIL) and class-incremental learning (Class-IL or CIL).
Formally, these three scenarios can be distinguished based on whether, at test time, task identity is provided and, if not, whether task identity must be inferred. This means that, in theory, any sequence of tasks could be performed according to all three scenarios. Fig.~\ref{fig:scenarios} illustrates what it means for the Split MNIST toy problem to be performed according to each scenario.
These three scenarios were initially described for task sequences with clear boundaries (i.e.,~task-based continual learning), but it has since been pointed out that they also generalize to task-free continual learning \citep{van2022three}. The key to generalizing these three scenarios is defining them based on how the non-stationary aspect of the data (i.e.,~the aspect of the data that changes over time, or the `task set') relates to the function or mapping that must be learned by the network.

Informally, with task-incremental learning a network must incrementally learn a set of distinct tasks. The use of `distinct tasks' here indicates that the network is always aware of which task it is presented with. Task identity might, for example, be known because it is explicitly provided, because it is clear from context which task must be performed, or because the inputs from different tasks are easily distinguishable from each other.
Thanks to the availability of task identity information, with task-incremental learning it is possible to use networks with task-specific components (e.g.,~a~separate output layer per task), or even to have a completely separate network per task~--~in which case there is no forgetting at all. Thus, simply preventing catastrophic forgetting is not difficult in this scenario. Instead, the challenge with task-incremental learning is to do \emph{better} (in terms of a trade-off between performance and resource efficiency) than the naive solution in which there is a separate network per task. To realize this, it is necessary to achieve a positive transfer between tasks by sharing learned representations across tasks.
An illustrative example of task-incremental learning is learning to play different musical instruments (e.g.,~first the piano, then the violin), since it should typically be clear which instrument must be played. 

Domain-incremental learning can be described as the structure of the problem to be learned being always the same, but the context or input-distribution changes (e.g., there are domain shifts).
In contrast to task-incremental learning, where the network always knows from which task an example is, with domain-incremental learning it is not necessarily clear to which domain a presented example belongs. As a result, in this scenario it is not possible to use networks with `domain'-specific components, unless the network also infers to which domain an example belongs (e.g.,~as done by \citealp{heald2021contextual,verma2021efficient}).
However, inferring domain identity can be challenging and is not necessarily the most effective way to solve a domain-incremental learning problem.
Illustrative examples of domain-incremental learning are learning to drive in various weather conditions or learning to classify objects under different lighting conditions (e.g.,~indoors versus outdoors).

Finally, class-incremental learning is the problem of incrementally learning to distinguish between an increasing number of objects or classes.
In the literature, this scenario is often implemented in a `task-based manner', meaning that there is a sequence of classification-based `tasks', with each task containing a distinct set of classes, and the goal is to learn to discriminate between the classes of all tasks. Let us illustrate this with an example. Imagine that a network is first presented with a task consisting of pianos and guitars, and later with one containing saxophones and violins. In the class-incremental learning scenario, after seeing both tasks, the network is expected to be able to discriminate between all four musical instruments (i.e.,~it should have learned a four-way classifier). On the other hand, if this same task sequence were performed according to the task-incremental learning scenario, the network would only be expected to be able to distinguish between instruments in the same task, but not between those from different tasks.
In this task-based setting, class-incremental learning can be decomposed into solving each individual task (or prediction within the task) and inference of task identity (or prediction across tasks) \citep{soutif2021importance,guo2023dealing}. Especially inferring task identity, which has links to out-of-distribution detection \citep{henning2021posterior,kim2022theoretical}, is often found to be particularly difficult. Beyond the task-based setting, this can be generalized by saying that a challenging aspect of class-incremental learning is learning to discriminate between classes that are not observed together.

\paragraph{Clarifying note.}
It is useful to point out that the distinction between task-, domain- and class-incremental learning is orthogonal to the distinction between task-based and task-free continual learning, in the sense that each of the three scenarios can occur in both a task-based and a task-free continual learning setting. In the literature, there is sometimes confusion about the difference between `task-based continual learning' and `task-incremental learning'. We therefore clarify that task-based continual learning refers to that training data change over time in a task-by-task manner (and with clear boundaries between tasks), while task-incremental learning signals that task identity is known to the network at test time.


\subsection{Evaluation}
\label{sec:evaluation}

With the rapid increase in interest in continual learning, numerous approaches have been proposed for evaluating and comparing different continual learning methods (e.g.,~\citealp{lopez2017gradient,diaz2018don,new2022lifelong,lange2023continual,kudithipudi2023design}). Typically, metrics for continual learning cover one of three areas: i)~performance, ii)~diagnostic analysis, and iii)~resource efficiency. 

With regards to evaluating performance in a continual learning setting, two important questions are: (a)~\emph{how} to evaluate it, and (b)~\emph{when} to evaluate it. Regarding the first question, continual learning performance is typically evaluated using the average of a certain performance metric (e.g.,~test accuracy in case of classification) over a family of tasks. Usually this family of tasks consists of all tasks trained on so far, or sometimes all tasks that will be trained on, and o the performance on all evaluated tasks is weighted equally. 
Regarding the question when to evaluate performance, one approach is to only do so at the end of training on all tasks.
Another common approach is to evaluate performance periodically throughout training by interleaving training and evaluation blocks. Performance can for example be evaluated after finishing training on each task, or after a fixed number of training steps.

A disadvantage of comparing continual learning methods only based on their average performance is that this does not provide much insight into how each method addresses continual learning. For example, a plastic model that achieves 0\%~accuracy on the first task and 100\%~accuracy on the second task, has the same average performance as a completely rigid model with 100\%~accuracy on the first task and 0\%~accuracy on the second. To provide greater insight into the dynamics of continual learning, there are several diagnostic metrics. 
One such diagnostic metric is learning accuracy. This measures well a model can learn from the current task, which, when compared to a baseline model, can provide a measure of plasticity. Another popular diagnostic metric is backward transfer, which measures how the performance on previous tasks changes when training on new ones. This metric provides insight into the degree of catastrophic forgetting that occurs and the stability of the model. Another diagnostic metric is forward transfer, which quantifies how much training on previous tasks improves the performance of a model on, or its ability to learn, future tasks.

A third set of metrics evaluates the resource efficiency of continual learning methods. 
One way to quantify resource efficiency is through the computational overhead and energy consumption of a method. The number of operations and the complexity are important factors when deploying continual learning methods to real-world problems, especially when targeting edge applications. It is important to note that although computational complexity is often only computed for the training phase, a portion of continual learning methods introduce additional complexity during inference, which can be measured as well and is not always identical to the training costs. Another key aspect of resource efficiency is sustainability, which refers to how quickly a model grows in terms of parameters, or in terms of memory for storing information about prior experiences (e.g.,~data samples, model copies).


\section{Continual Learning Approaches}
\label{sec:approaches}

In this section we focus on approaches that have been proposed to address the continual learning problem. We do not provide an extensive review of individual continual learning methods (for this, we refer to \citealp{belouadah2020comprehensive,delange2021continual,masana2020class,wang2023comprehensive}), but rather we discuss the main computational strategies that underlie these methods. Individual continual learning methods often combine multiple of these strategies.   


\subsection{Replay}

\begin{figure}[t]
  \centering
  \includegraphics[width=0.76\linewidth]{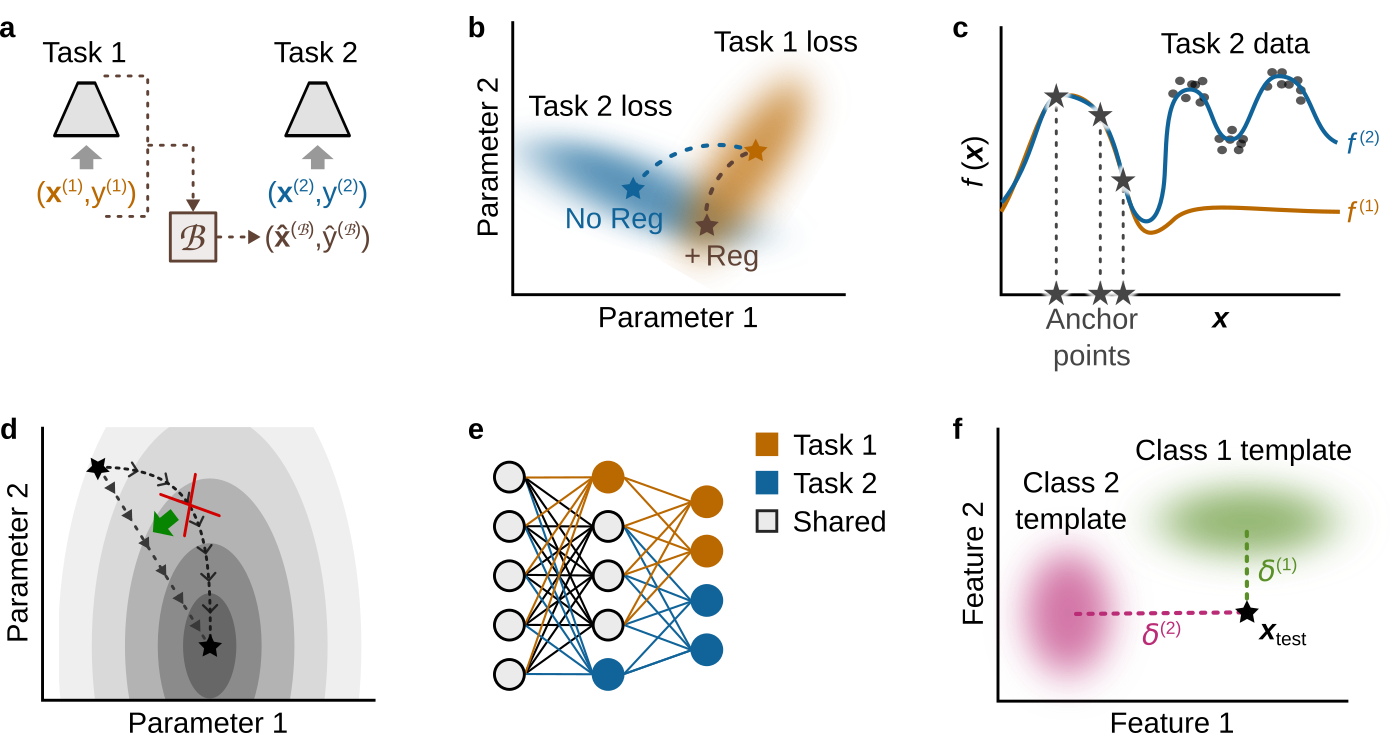}
  \caption{\textbf{Overview of different computational approaches for continual learning.}
  \textbf{a},~Replay. Current training data are complemented with replayed data representative of previously seen data.
  \textbf{b},~Parameter regularization. A penalty term is added to the loss to encourage parameters important for past tasks not to change too much during continued training.
  \textbf{c},~Functional regularization. A penalty term is added to the loss to encourage the input-output mapping of the network not to change too much at a particular set of inputs (the anchor points) during continued training.
  \textbf{d},~Optimization-based approaches. Modifications are made to the way a given loss function is optimized.
  \textbf{e},~Context-dependent processing. Depending on which task or context is presented, only specific parts of the network are used. 
  \textbf{f},~Template-based classification. For each class a template is learned, and test samples are classified based on which class template is the most suitable for them.
  Except for panel~d, this figure has been modified from Fig.~3 of \citet{van2022three}.}
  \label{fig:strategies}
\end{figure}

Perhaps the most widely used approach for continual learning is replay. The idea behind replay, which is also referred to as rehearsal, is to approximate interleaved learning by complementing the training data of the current task or experience with data that are representative of previous ones (Fig.~\ref{fig:strategies}a).
Replay has close links to neuroscience. In the brain, the re-occurence of neuronal activity patterns that represent previous experiences is believed to be important for the stabilization and consolidation of new memories \citep{wilson1994reactivation,rasch2007maintaining}.

In the deep learning literature, a common way to add replay to a neural network is to store previously seen data in a memory buffer and revisit them later on. Such `experience replay' can substantially speed up training in reinforcement learning \citep{lin1992self,mnih2015human}, and it has proven to be an effective way to reduce catastrophic forgetting in continual learning \citep{robins1995catastrophic,rolnick2019experience,chaudhry2019tiny,buzzega2020dark}. A common assumption in benchmarks for continual learning is that only a limited amount of data can be stored (but see~\citealp{prabhu2023computationally,verwimp2023continual} for recent perspectives on this assumption), and a research question that has received a lot of attention is how to best pick the samples to store in the memory buffer \citep{rebuffi2017icarl,chaudhry2019tiny,aljundi2019gradient,mundt2023wholistic}.

Instead of explicitly storing past observations, it is also possible to learn a generative model, which can then be used to generate the data to be replayed \citep{mocanu2016online,shin2017continual,wu2018memory,rao2019continual,cong2020gan,khan2023looking}. An issue with such `generative replay' is that it can be difficult to train generative models of decent quality, especially in an incremental setting or when data are complex \citep{aljundi2019online,lesort2019generative}. This issue can sometimes be alleviated by replaying latent features instead of raw inputs \citep{liu2020generative,pellegrini2020latent,ostapenko2022continual}, but this `internal replay' approach needs some form of pretraining to work well.

When replay is used to train a deep neural network, the loss on the replayed data~($\ell_{\text{replay}}$) is usually added to the loss on the current data~($\ell_{\text{current}}$), possibly with some weighting applied, and the objective is to optimize the combined loss:
\begin{equation}
\ell_{\text{total}}(\boldsymbol{\theta}) = \ell_{\text{current}}(\boldsymbol{\theta}) + \ell_{\text{replay}}(\boldsymbol{\theta})
\label{eq:replay}
\end{equation}
with $\boldsymbol{\theta}$ the parameters of the network that can be updated during training. 

An alternative way to use the loss on the replayed data is to construct one or more inequality constraints that should be respected when optimizing the loss on the current data \citep{lopez2017gradient,chaudhry2019efficient}. Although this strategy, which is referred to as `gradient episodic memory', is sometimes categorized in the literature under replay, we consider it an optimization-based approach and discuss it in subsection~\ref{sec:optimization_based}.

Especially for large and complex continual learning problems, replay seems to play an important and perhaps necessary role. However, as replay involves constantly retraining on past data, an important concern with this approach is the potentially high computational cost. Luckily, there is evidence that it is not necessary to fully (re)train on all past tasks whenever a new task is learned. One reason is that learning something new is typically more demanding than preventing its forgetting once it has been learned, which explains why it can be sufficient to replay only relatively small amounts of data \citep{vandeven2020brain}. In the cognitive science literature, it has further been suggested that it might only be needed to replay old data that are similar to the new data \citep{mcclelland2020integration}. The intuition for this is that interference between unrelated items should be small anyway (for example because such items are stored in separate parts of the network). This idea has inspired a series of continual learning studies asking how to adaptively select which data to replay given the currently observed data \citep{riemer2019learning,aljundi2019online,klasson2023learn,krawczyk2023adiabatic}.
Other open questions with regards to replay are how to store and compress data, in what format to replay data (e.g.,~raw data or intermediate features), and how to best integrate replay with other methods.


\subsection{Parameter Regularization}
\label{sec:param_reg}
Another popular approach for continual learning is parameter regularization. When a new task is learned, parameter regularization discourages large changes to parameters of the network that are thought to be important for previous tasks~(Fig.~\ref{fig:strategies}b). From a neuroscience perspective, this approach can be linked to metaplasticity \citep{abraham2008metaplasticity}, as it can be interpreted as equipping the network parameters with an internal state that modulates their level of plasticity \citep{zenke2017improved}. Another motivation for parameter regularization comes from a Bayesian perspective, as instances of this approach can often be expressed or interpreted as performing sequential approximate Bayesian inference on the parameters of a neural network \citep{kirkpatrick2017overcoming,nguyen2018variational,farquhar2019unifying}.

We define parameter regularization as adding a regularization term to the loss function to penalize changes to the network's parameters $\boldsymbol{\theta}$, whereby the applied penalty is usually weighted by an estimate of how important parameters are for previously learned tasks (e.g., \citealp{kirkpatrick2017overcoming,zenke2017improved,aljundi2018memory,schwarz2018progress,ritter2018online}):
\begin{equation}
\ell_{\text{total}}(\boldsymbol{\theta}) = \ell_{\text{current}}(\boldsymbol{\theta}) + \norm{\boldsymbol{\theta}-\boldsymbol{\theta}^*}_\Sigma
\label{eq:param_reg}
\end{equation}
with $\boldsymbol{\theta}^*$ the value of the parameters relative to which changes are penalized (often this is the value of $\boldsymbol{\theta}$ at the end of the previous task), $\Sigma$ an estimate of how important the parameters are for past tasks and $\norm{.}_\Sigma$ a weighted norm. The most common choice is a weighted $L^2$-norm, in which case the regularization term becomes $\frac{1}{2}\left(\boldsymbol{\theta}-\boldsymbol{\theta}^*\right)^T\Sigma\left(\boldsymbol{\theta}-\boldsymbol{\theta}^*\right)$.
We note that it is also possible to use estimates of the importance of parameters for past tasks in other ways, for example to reduce the learning rate for relatively important parameters (e.g.,~\citealp{ozgun2020importance}) or to perform gradient projection during optimization (e.g.,~\citealp{kao2021natural}). Although in the literature these approaches are sometimes categorized under parameter regularization, we consider them optimization-based approaches and discuss them in subsection~\ref{sec:optimization_based}.

A pivotal aspect of parameter regularization is estimating the importance of the network's parameters for past tasks. An often-used method is to leverage the Fisher Information matrix \citep{kirkpatrick2017overcoming}, which, under certain assumptions, indicates how a small change to the parameters would impact the loss. The Fisher Information is typically approximated with a diagonal matrix, thus assuming independence among all parameters. However, this assumption can be relaxed, for example by instead using a Kronecker-factored approximation \citep{martens2015optimizing,ritter2018online,kao2021natural}. An important disadvantage of the Fisher Information is that it can be costly to compute. Several other parameter regularization methods instead estimate parameter importance online throughout training, which often incurs substantially lower computational costs \citep{zenke2017improved,aljundi2018memory}.

Although parameter regularization methods have shown success in task- and domain-incremental learning problems, they often struggle to learn inter-task boundaries in class-incremental learning scenarios \citep{lesort2019regularization,van2022three,kessler2023sequential}. 


\subsection{Functional Regularization}
\label{sec:func_reg}

An inherent difficulty with parameter regularization is that correctly estimating the importance of parameters for past tasks is very hard, which is due to the complex relation between the behaviour of a deep neural network and its parameters.
Instead of operating in the parameter space, a more effective approach might be applying regularization in the function space of a neural network \citep{benjamin2019measuring,pan2020continual,titsias2020functional}.
The goal of such functional regularization is to prevent large changes to a network's input-output mapping $f_{\boldsymbol{\theta}}$ at a set of specific inputs, which are termed `anchor points' (Fig.~\ref{fig:strategies}c). Similar to parameter regularization, functional regularization can be expressed as adding a penalty term to the loss function:
\begin{equation}
\ell_{\text{total}}(\boldsymbol{\theta}) = \ell_{\text{current}}(\boldsymbol{\theta}) + \left<f_{\boldsymbol{\theta}}, f_{\boldsymbol{\theta}^*}\right>_{\mathcal{A}}
\label{eq:func_reg}
\end{equation}
with $f_{\boldsymbol{\theta}^*}$ the input-output mapping relative to which changes are penalized (often this is the input-output mapping of the network at the end of the previous task) and $\mathcal{A}$ the set of anchor points where the divergence between $f_{\boldsymbol{\theta}}$ and $f_{\boldsymbol{\theta}^*}$ is evaluated.

There are different ways in which the divergence between $f_{\boldsymbol{\theta}}$ and $f_{\boldsymbol{\theta}^*}$ can be measured. For classification-based problems, following \citet{li2017learning}, a popular choice is to use the knowledge distillation loss proposed by \citet{hinton2015distilling}, which involves the cross entropy between temperature-scaled logits of both networks. However, the divergence between $f_{\boldsymbol{\theta}}$ and $f_{\boldsymbol{\theta}^*}$ does not need to be measured at the output level. In recent years, functional regularization is increasingly applied at various levels of the representation of neural networks, in which case it is also called feature distillation \citep{heo2019comprehensive,douillard2020podnet,gomezvilla2022continually,roy2023subspace}.

Another important aspect of functional regularization is the selection of anchor points. For relatively simple problems, \citet{robins1995catastrophic} demonstrated that functional regularization with random patterns as anchor points, which he called `pseudorehearsal', can already work reasonably well. However, for more complex problems, it is important that the set of anchor points is representative of the inputs from previous tasks. A naive solution would be to use all inputs that have been seen so far as anchor points, but this requires storing those inputs and functional regularization with a large number of anchor points can incur high computational costs. An option that does not involve storing past samples is using the currently observed inputs as anchor points \citep{li2017learning}. This approach, which is known as `learning without forgetting', tends to work well when the inputs from different tasks have similar structure (e.g.,~as is the case with tasks that all consist of natural images), but in general there is no guarantee that the current inputs are suitable anchor points for previous tasks. Another option is to use as anchor points a small number of strategically selected inputs from previous tasks that represent those tasks well. One way to select such representative inputs is by formulating neural networks as Gaussian Processes \citep{khan2019approximate}, as this allows for summarizing the input distributions of previous tasks with inducing points \citep{titsias2020functional} or memorable inputs \citep{pan2020continual}. How to optimally select the anchor points for functional regularization is still largely an open question.

Functional regularization is closely related to replay. In fact, functional regularization can be interpreted as a form of replay, whereby the replayed data consist of the anchor points labelled with the predictions for those points made by (a previous version of) the network itself. The key difference between both approaches is that with replay inputs \emph{and} targets of past tasks are stored externally to the network (e.g., in a memory buffer or in the form of a separate generative model), while with functional regularization only inputs of past tasks are stored externally.
With functional regularization, the input-output mapping of past tasks is therefore only stored internally in the network itself (or in its copies), while with replay at least part of this mapping is stored externally.
However, in the continual learning literature, the distinction between replay and functional regularization is sometimes blurred. For example, with generative replay there is often only a generative model for the input distribution, and the generated inputs that are replayed are labelled based on predictions made for them by a previous version of the network (e.g.,~\citealp{shin2017continual}). Consequently, despite its name, such generative replay is actually a form of functional regularization. Moreover, the replay of stored data from a memory buffer has also been combined with distillation (e.g.,~\citealp{buzzega2020dark,boschini2023class}), making those instances a form of functional regularization as well.


\subsection{Optimization-based Approaches}
\label{sec:optimization_based}

The three approaches discussed so far~--~replay, parameter regularization and functional regularization~--~operate by making changes to the loss function that is optimized (as shown by equations~\ref{eq:replay}-\ref{eq:func_reg}). An alternative approach to continual learning is to change \emph{how} the loss function is optimized~(Fig.~\ref{fig:strategies}d).
The standard optimization routines that are used in deep learning, such as stochastic gradient descent~(SGD) and its variants (e.g.,~AdaGrad, \citealp{duchi2011adaptive}; Adam, \citealp{kingma2014adam}), have been developed for stationary settings. In non-stationary settings, there are typically no guarantees for their behavior, yet these standard optimization routines are the default choice in most work on continual learning. However, in the last few years there has been an increasing attention in the continual learning literature for the role of optimization, and there have been several attempts to develop novel optimization routines specific for continual learning. It has even been argued that an wholistic solution for continual learning must consist of both changes to the loss function and changes to how that loss function is optimized \citep{hess2023two}.

Already about six years ago it was empirically found that the type of optimizer (e.g., plain SGD, AdaGrad or Adam) can have a large effect on continual learning performance \citep{hsu2018re}. Later work explored a wider range of factors that influence optimization (e.g., learning rate, batch size) and concluded that the best optimization routines for continual learning are the ones that tend to find wider or flatter minima \citep{mirzadeh2020understanding}. This can be explained because with a wider minimum, larger changes to the parameters are needed to `get out of the minimum'. This insight has motivated several continual learning works to modify optimization routines to encourage finding such wider minima \citep{deng2021flattening,yang2023data,tran2023sharpness}.

A popular way to control how a given loss function is optimized is by using adaptive learning rates. For example, one strategy might be to reduce the learning rate for either parameters or units that are estimated to be important for past tasks \citep{ahn2019uncertainty,jung2020continual,ozgun2020importance,paik2020overcoming,laborieux2021synaptic,soures2021tacos,malviya2022tag}. This approach is related to parameter regularization, but it is different because the use of adaptive learning rates does not change the loss function, while parameter regularization does. Similar to parameter regularization, the use of adaptive learning rates can be related to the neuroscience concept of metaplasticity \citep{abraham2008metaplasticity}.
A different metaplasticity-inspired way to control the optimization trajectory is through probabilistic parameter updates \citep{zohora2020metaplasticity,SS} rather than adjusting the learning rate.

Another optimization-based tool that has been explored in continual learning is gradient projection. With gradient projection, rather than basing parameter updates on the original gradient~$g=\nabla_{\boldsymbol{\theta}}\ell\left(\boldsymbol{\theta}\right)$, they are based on a projected version~$\bar{g}$ of that gradient.
A first popular approach is `orthogonal gradient projection' \citep{zeng2019continual,farajtabar2020orthogonal,saha2021gradient}. To restrict parameter updates to directions that do not interfere with the performance on old tasks, this approach projects gradients to the subspace orthogonal to the gradients of old tasks.
A less restrictive version of this approach instead projects gradients using the inverse of the Fisher Information as projector matrix \citep{kao2021natural}.
Another gradient projection-based approach is `gradient episodic memory' \citep{lopez2017gradient,chaudhry2019efficient}. The projection mechanism of this approach is derived from a constrained optimization problem which aims to optimize the loss on a new task without increasing the loss on old tasks.

Another method that falls into the category of optimization-based techniques is based on the biological process of synaptic consolidation, which consolidates information within an individual synapse over multiple timescales \citep{sossin_2008_Molecular,morris_2003_Longterm}. 
Such synaptic consolidation can be modelled with a complex synaptic model consisting of a rapidly adapting weight~$\boldsymbol{\theta}_f$ and a slowly evolving weight~$\boldsymbol{\theta}_s$. The rapidly adapting weight,~$\boldsymbol{\theta}_f$, is driven by a combination of the loss function and a regularization-like function that attempts to prevent divergence from~$\boldsymbol{\theta}_s$. The main distinction between parameter regularization and this approach lies in the dynamic nature of~$\boldsymbol{\theta}_s$, which slowly evolves to consolidate the information learned by~$\boldsymbol{\theta}_f$. Several studies have used this approach to reduce catastrophic forgetting in continual learning \citep{zenke2015diverse,leimer_2019_Synaptic,soures2021tacos}.


\subsection{Context-dependent Processing}
\label{sec:context_dependent}

Another popular approach for continual learning is context-dependent processing. The idea behind this approach is to use certain parts of the network only for specific tasks or contexts, in order to reduce the interference that can occur between them~(Fig.~\ref{fig:strategies}e). It is worth noting that when taken to the extreme, this approach corresponds to having a completely separate network per task or context. In this case, there would be no interference or forgetting at all, but there would also no longer be any possibility of positive transfer between tasks or contexts.
It could therefore be argued that continual learning methods should aim to only segregate information that is unrelated to each other (as there is likely no positive transfer to be gained between them anyway), while storing related information in the same part of the network.

A widespread example of context-dependent processing in continual learning is the use of a separate linear output layer for each task to be learned. The use of such a `multi-headed output layer' has become the default setup for task-incremental learning experiments. But there are other ways in which context-dependent processing is used as well.

One other way to induce context-dependent processing in a neural network is by gating either its units or its parameters in a different way for each task that must be learned. It is possible to specify such task-specific gates \emph{a priori} and randomly \citep{masse2018alleviating}, but it is also possible to learn them, for example using gradient descent \citep{serra2018overcoming}, Hebbian plasticity \citep{flesch2023modelling} or evolutionary algorithms \citep{ellefsen2015neural,fernando2017pathnet}.
The use of task- or context-specific gates has been linked to the concept of neuromodulation in the brain, which refers to chemical signals that locally modify the way inputs are processed \citep{marder2002cellular}.

Another way to realize context-dependent processing is by periodically adding new components, and thus dynamically expanding the network \citep{zhou2012online,terekhov2015knowledge,draelos2017neurogenesis}. Expanding the network can be based on when extra capacity is needed \citep{hung2019compacting,mitchell2023self}, but in many cases new components are simply added whenever a new task must be learned \citep{rusu2016progressive,yoon2018lifelong}. Often, this approach is combined with freezing parts of the network after a task has been learned. To control the growth of the network, many model expansion methods make use of sparse training (e.g.,~prompt learning or the use of adaptors; \citealp{wang2022learning,gao2023lae}) or pruning techniques \citep{golkar2019continual,pandit2020relational,yan2021dynamically}.
The approach of dynamically expanding the network can be linked to the process of neurogenesis, which refers to the addition of neurons to certain brain regions throughout life \citep{aimone2014regulation}. It has indeed been hypothesized that neurogenesis plays a role in the brain's ability to mitigate catastrophic forgetting \citep{wiskott2006functional}.

An important assumption that often underlies the use of context-dependent processing in continual learning is that the context (or task) is always clear to the network. As a result of this, when used by itself, this approach is limited to task-incremental learning type of problems. To be applicable to domain- or class-incremental learning problems, context-dependent processing must be combined with an algorithm for context identification (e.g.,~as done by \citealp{aljundi2017expert,von2019continual,wortsman2020supermasks,henning2021posterior,verma2021efficient,heald2021contextual}). In this regard it is important to realize that context identification itself is a class-incremental learning problem as well, as it consists of distinguishing categories (in this case `contexts') that are not observed together. Inferring context information has indeed proven to be a difficult problem, and presents a promising avenue for future research.


\subsection{Template-based Classification}
\label{sec:template_based}

An approach to class-incremental learning that is often used in continual learning is template-based classification. With this approach, a `class template' is learned for every class, and classification is performed based on which class template is closest or most suitable for the sample to be classified~(Fig.~\ref{fig:strategies}f). In this description, a class template can be thought of as a representation or a model of that particular class.
In the context of class-incremental learning, an important advantage of template-based classification is that it avoids the need to make comparisons between classes \emph{during training}. Standard softmax-based classifiers have to learn decision boundaries between all classes during their training, but this is challenging with class-incremental learning because not all classes are observed together. Template-based classifiers instead only have to learn a template per class during their training, and the comparison between classes is deferred to test time.
Importantly, while the original problem is a class-incremental learning problem, learning these class templates is a task-incremental learning problem, whereby each `task' is to learn a template for a specific class. This means that with this approach it is possible to use `template-specific components', or other context-dependent processing approaches.

A popular way to implement template-based classification is to use `prototypes' as class templates. The use of such prototypes has roots in cognitive science, where it is an influential model for how humans make categorization decisions \citep{nosofsky1986attention}.
In deep learning, a prototype is often taken to be the mean vector of examples from a class in an embedding space defined by a neural network \citep{snell2017prototypical,yang2018robust}.
Classification is then done by putting samples through the embedding network and assigning them to the class of the prototype they are closest to.
When a suitable pretrained embedding network is available and it is kept frozen throughout training, storing raw data is not needed and prototype-based classification only requires storing a single prototype per class (e.g.,~\citealp{hayes2020lifelong}).
If the embedding network requires updates, for example to better distinguish new classes, one option is to store a few well-chosen examples per class to update the prototypes following changes to the embedding network \citep{rebuffi2017icarl, de2020continual}.
Alternatively, prototype drift can be addressed without storing past examples by estimating and correcting drift based on the observed drift for data from the current task \citep{yu2020semantic,wei2021incremental}.

Another example of template-based classification that has shown promise for class-incremental learning is generative classification \citep{vandeven2021class,banayeeanzade2021generative}. With generative classification, the class templates that are learned are generative models, and their suitability for a sample to be classified is computed as the sample's likelihood under each of the generative models. 
An advantage of generative classification is that it does not require storing samples, but a drawback is that learning decent generative models can be challenging, especially in the case of limited training data or complex distributions. Moreover, inference (i.e.,~making classification decisions) can be computationally costly as it requires computing likelihoods under the generative model of each class.
Several more lightweight and efficient alternatives for generative classification have been proposed. One option is to train an energy-based model and compute energy values per class or context instead of likelihoods \citep{li2022energy,joseph2022energy}. Another option is to train class-specific models to replicate the outputs of a frozen random network and perform classification based on each model's prediction error \citep{zajkac2024prediction}.


\section{Continual Learning in Deep Learning versus in Cognitive Science}
\label{sec:humans}

So far, this book chapter has mostly reviewed the deep learning literature on continual learning. In this section, we briefly ask how continual learning is studied in the cognitive science literature. When it comes to continual learning, the goals of deep learning and cognitive science are different, but clearly related. 
Deep learning aims to engineer artificial neural networks so that they can learn continually, while cognitive science is concerned with understanding how this already present skill is implemented by the brain.

One central aspect of continual learning that is widely studied in cognitive science is forgetting. Forgetting refers to the loss or decay of previously acquired information and can result from various factors in biological organisms \citep{hardt2013decay,davis2017biology}. There exist passive mechanisms, such as decay over time through natural aging and transient forgetting in which forgetting is often reversible, and there are active mechanisms of forgetting, such as intentional forgetting, retrieval-induced forgetting and interference-based forgetting.
An example of interference-based forgetting is retroactive interference, where acquisition of new memories during consolidation leads to forgetting of old ones \citep{wixted2004psychology,alves2017retroactive}, which is probably most akin to the forgetting that happens when deep neural networks are continually trained. Studies into how biological systems mitigate this retroactive interference have already inspired many approaches for continual learning with deep neural networks (e.g.,~\citealp{kaplanis2018continual,masse2018alleviating,tadros2022sleep,kudithipudi2022biological,arani2022learning,wang2023incorporating,jeeveswaran2023birt}), and it seems likely that deep learning can still gain further benefits from the rich characterization and understanding of forgetting that has been accrued in the cognitive science literature. For example, something not often acknowledged in deep learning is that with bounded resources, forgetting could be beneficial or even necessary. Indeed, some deep learning applications might benefit from controlled or intentional forgetting, similar to how forgetting is thought to be important for the brain's ability to continue learning new information (e.g.,~\citealp{norby2015forget,richards2017persistence,bjork2019forgetting}).

While forgetting is well studied in cognitive science, other aspects of continual learning (cf.~subsection~\ref{sec:other_features}) appear to receive less attention; or at least the way in which these other aspects are studied in deep learning is different and mostly unconnected from the way they are studied in cognitive science. 
However, there are some efforts to bridge this gap between both fields. In particular, \cite{flesch2018comparing} designed several experiments to compare the continual learning performance of humans and artificial neural networks. They demonstrate that, unlike deep learning models that suffer from blocked training, humans actually perform better with blocked training (and worse with interleaved training). In another work, \citet{flesch2023modelling} show how, through the incorporation of different mechanisms, the learning dynamics of artificial neural networks can be made to more closely match those observed in humans.
Establishing more such connections between cognitive science and deep learning in the context of continual learning, especially if they go beyond just forgetting, might substantially benefit both fields.


\section{Conclusion}
\label{sec:conclusion}

In this book chapter we have reviewed the challenges of continual learning with deep neural networks, which is a topic of great interest in the artificial intelligence field. The current inept ability of deep learning models to continually learn from a stream of incoming data is a major obstacle to the development of truly intelligent artificial agents that can accumulate knowledge by incrementally learning from their experiences.

Catastrophic forgetting is the most well-known issue in continual learning. However, in addition to overcoming catastrophic forgetting, a complete solution to continual learning requires the development of models capable of quickly adapting to new situations, exploiting similarities between tasks, operating in a task agnostic manner, being tolerant to noise, and using resources in an efficient and sustainable manner. Continual learning is thus not a unitary problem, and this is further illustrated by the observation that the main challenges of continual learning can differ substantially depending on how exactly the problem is set up. To fully capture the gradations and complexities of continual learning, a variety of settings and benchmarks needs to be considered. Among these are task-, domain- and class-incremental learning, as well as task-based and task-free continual learning. Similarly, comprehensively evaluating continual learning approaches cannot be done using a single metric; in this book chapter we have discussed a variety of continual learning metrics covering performance, diagnostics and resource efficiency.

We have further reviewed six computational approaches for continual learning. Replay, parameter regularization and functional regularization operate by making changes to the loss function, while optimization-based approaches change the way in which a given loss function is optimized. Context-dependent processing distributes computations across the network based on contextual or task-specific information, and template-based classification enables distinguishing objects or categories that are not observed together without having to directly learn discriminative boundaries.

As deep learning navigates the intricate challenges posed by non-stationary environments and catastrophic forgetting, it becomes evident that continual learning is not just a technical necessity, but a philosophical shift in the approach to artificial intelligence. Approaching this from an interdisciplinary collaboration, drawing inspiration from neuroscience, cognitive science, and psychology to imbue continual learning into artificial systems, seems to hold promise. Similarly, but in the reverse direction, insights gained in the deep learning field while doing so can help to unravel the computational principles that underlie the cognitive skill of continual learning in the brain.



\subsection*{Acknowledgements}
This effort is partially supported by the NSF EFRI BRAID Award \#2317706 and the NSF PARTNER AI Institute NAIAD Award \#2332744, as well as by funding from the European Union under Horizon Europe (Marie Skłodowska-Curie fellowship, grant agreement No.~101067759).

\newpage
\bibliographystyle{agsm}
\bibliography{References.bib}


\end{document}